\pdfoutput=1

\documentclass[11pt]{article}

\usepackage[]{acl}

\usepackage{times}
\usepackage{latexsym}

\usepackage[T1]{fontenc}

\usepackage[utf8]{inputenc}

\usepackage{microtype}

\usepackage{amsfonts,amsmath,bm,xspace}
\usepackage{graphicx}
\usepackage{subcaption}
\usepackage{booktabs}
\usepackage{multicol,multirow}
\usepackage{paralist}
\usepackage{array,hhline}
\usepackage{bbm}
\usepackage{diagbox}
\usepackage{wrapfig}
\usepackage{amssymb}
\usepackage{amsthm} 
\usepackage{algorithm2e}

\DeclareMathOperator*{\argmax}{arg\,max}
\DeclareMathOperator*{\argmin}{arg\,min}

\newcommand{\x}{X}
\newcommand{\y}{Y}
\newcommand{\z}{\bm z}

\newcommand{\N}{N}
\newcommand{\F}{\mathbb{F}}
\newcommand{\Q}{\bm q}

\newcommand{\uno}[1]{\overline{#1_\text{obs}}}
\newcommand{\obs}[1]{#1_\text{obs}}

\newcommand{\hit}{\checkmark}
\newcommand{\method}{\textit{latent-}GLAT\xspace}

\title{\method: Glancing at Latent Variables for Parallel Text Generation}

\author{Yu Bao$^{\spadesuit\diamondsuit}$,\ Hao Zhou$^{\clubsuit}$,\ Shujian Huang$^{\spadesuit\diamondsuit}$\footnotemark[1]\ , Dongqi Wang$^{\spadesuit\diamondsuit}$\\
{\bf Lihua Qian$^{\clubsuit}$,\ Xinyu Dai$^{\spadesuit\diamondsuit}$,\ Jiajun Chen$^{\spadesuit\diamondsuit}$,\ Lei Li$^{\heartsuit}$\footnotemark[2]} \\
        $^{\spadesuit}$National Key Laboratory for Novel Software Technology, Nanjing University, China \\
        $^{\diamondsuit}$Collaborative Innovation Center of Novel Software Technology and Industrialization,  China\\
        $^{\clubsuit}$Bytedance AI Lab, China \quad $^{\heartsuit}$University of California Santa Barbara, USA\\
        \texttt{\{baoy,wangdq\}@smail.nju.edu.cn}
        \\
        \texttt{\{zhouhao.nlp,qianlihua\}@bytedance.com}
        \\
        \texttt{\{huangsj,daixinyu,chenjj\}@nju.edu.cn}
        \quad \texttt{leili@cs.ucsb.edu}
        }

\begin{document}
\maketitle
\renewcommand{\thefootnote}{\fnsymbol{footnote}}

\footnotetext[1]{Shujian Huang is the corresponding author.}
\footnotetext[2]{Work is done while at ByteDance AI Lab.}
\begin{abstract}
Recently, parallel text generation has received widespread attention due to its success in generation efficiency.  
Although many advanced techniques are proposed to improve its generation quality, they still need the help of an autoregressive model for training to overcome the one-to-many multi-modal phenomenon in the dataset, limiting their applications.
In this paper, we propose $\textit{latent}$-GLAT, which employs the discrete latent variables to capture word categorical information and invoke an advanced curriculum learning technique, alleviating the multi-modality problem.
Experiment results show that our method outperforms strong baselines without the help of an autoregressive model, which further broadens the application scenarios of the parallel decoding paradigm.
\footnotemark[3]
\end{abstract}
\footnotetext[3]{The implementation of \method will be released at \url{https://github.com/baoy-nlp/Latent-GLAT}.}
\renewcommand{\thefootnote}{\arabic{footnote}}
\setcounter{footnote}{0}
\section{Introduction}
Non-autoregressive Transformer~\cite[NAT,][]{nat} introduce a parallel decoding paradigm with higher decoding efficiency~(> $10\times$) than autoregressive models~\cite{attn_seq_to_seq,cnn_seq,transformer}.
Unlike autoregressive models, NAT models impose conditional independence assumptions in words to support parallel decoding of sentences during inference.
It attracts many researchers to explore NAT in machine translation~\cite{nat,iter_nat,lt} and text-to-speech tasks~\cite{chen2019listen,peng2020non}.

Amount of researchers devoted themselves to improve the NATs' inferior generation quality.
Such as modeling word inter-dependencies by curriculum learning~\cite{guo2020fine,liutask} or iterative refinements mechanism~\cite{cmlm,jm_nat}, introducing latent variables to decompose target sentences and serve as the springboard for decoding~\cite{lv_nar,flowseq,cnat}, and introduce inductive bias for models' training~\cite{imitate_nat,hint_nat}.
The most successful method is the glancing transformer~\cite[GLAT,][]{glat}, which trains the NAT model by sampling partial target words as inputs to predict the remaining target words, explicitly building dependencies between the observed and unobserved words. 
\citet{qian2021volctrans} employ GLAT to achieve impressive results on the
translation task of WMT21\footnote{\url{http://statmt.org/wmt21/}}, even outperforming many strong autoregressive translation systems in BLEU score~\cite{bleu}.

Although existing NAT models achieve competitive results compared to autoregressive models in translation tasks, it is not negligible that they still need the help of an autoregressive Transformer~\cite[AT,][]{transformer} as a teacher for training, i.e., sequence-level knowledge distillation~\citep{kim2016sequence}.
A well-recognized explanation is a \textit{multi-modality problem}~\cite{zhou2019understanding,emnat}: each input may have multiple valid outputs in datasets, which will prevent NAT models from learning to organize consistent outputs.
Training with the outputs of an AT can directly bypass the multi-modal phenomenon in the dataset, effectively improving the models' performances.
However, training NAT models by knowledge distillation are limited. 
First, it needs to train an extra AT model, which inevitably enlarges the training cost. 
Second, it is hard to promise that the teacher (or AT) model can be accurate enough in all text generation settings, which will become the bottleneck for its student NAT model. Therefore, training a model from scratch without the help of an AT model is still an open and interesting problem.

In this paper, we propose~\method, which can directly learn from the raw dataset.
It alleviates the multi-modality problem following a divide-and-conquer spirit, introducing a small set of discrete latent variables to capture the target word categorical information and divide the origin goal into latent variables modeling and sentence reconstruction.
First, the categorical information may have fewer multi-modality phenomena than the original words, thus can be learned directly without the help of knowledge distillation.
Second, the word categorical information is informativeness to the sentence reconstruction. We can extend glancing training with these discrete latent variables for modeling the sentence, encouraging the model to build dependencies on word categorical information rather than words, which works more robustly.

Experiment results on WMT14, Quora, and DailyDialog datasets show that \method achieves remarkable improvements over several strong baselines, verifying the effectiveness of \method.
More impressively, \method even outperforms autoregressive models in Quora and DailyDialog datasets, further validating our motivation for removing knowledge distillation.
In-depth analyses indicate that the introduced discrete latent variables are helpful to alleviate the multi-modality problem and are necessary for performance improvement.

\section{Background}\label{s:background}
For a sequence-to-sequence task of predicting sequence $\y=(y_1, y_2, \cdots, y_m)$ given its input sequence $\x=(x_1, x_2, \cdots, x_n)$, 
the classical autoregressively factorization decomposes the $p(\y|\x)$ with a series of conditional probability: 
\begin{equation}
    p_\text{AT}(\y|\x)= \prod_{t=1}^{m}p(y_{t}|y_{<t},\x),
    \label{eqn:at}
\end{equation}
where $y_{<t}=(y_1, y_2, \cdots, y_{t-1})$.

Although such factorization achieved great success in previous studies~\cite{attn_seq_to_seq,cnn_seq,transformer}, they predict each word\footnote{We use BPE segmentation in our experiments, and they are strictly tokens. For clarity, we use words and tokens interchangeably in the paper.} based on the prefix words, which may suffer from the issues of error accumulation and slow decoding during inference.

\paragraph{Non-autoregressive Transformer.}
To tackle the above problems, \citet{nat} firstly propose non-autoregressive Transformer (NAT), introducing a non-autoregressive factorization as:
\begin{equation}
    p_\text{NAT}(\y|\x) = \prod_{t=1}^{m}p(y_{t}|\x),
    \label{eqn:nat}
\end{equation}
where each word $y_t$ are modeled independently.
During inference, the NAT model can decode the word simultaneously by $\argmax_{y_t}p(y_t|\x)$ for each $y_t$, remarkably improving the efficiency~(15$\times$ speedups to an autoregressive Transformer).

However, the independence assumption may prevent the NAT model from leveraging the inherent word dependencies to organize consistent outputs. 
Due to this,the efficiency improvements of NAT are at the cost of its quality, e.g., the performance degradation by more than 10.0 BLEU~\cite{bleu} points in machine translation tasks~\cite{nat}.
Besides, recent studies~\cite{zhou2019understanding,emnat} point out that the multi-modality phenomenon in the dataset aggravates the challenge of NAT models.


\paragraph{Glancing Transformer.}
To mitigate the issue of missing word dependency in NAT models, \citet{glat} propose Glancing Transformer (GLAT), introducing glancing training (GLT) and sampling partial target tokens for training NAT:
\begin{equation}
\label{eqn:glt} 
\begin{split}
\mathcal{L}_\text{GLAT} &= -\log p(\overline{\obs{\y}} | \obs{\y},\x) \\
                       &= -\sum_{y_i\in \overline{\obs{\y}}}\log p(y_i|\obs{\y},\x), 
\end{split}
\end{equation}
where $\obs{\y}$ is the partial target tokens, and $\overline{\obs{\y}}$ is its complements set.
It progressively decreases the sampling ratio and obtains better performances in machine translation tasks.

Nevertheless, we find that GLAT in experiments still has a multi-modality problem\footnote{We include details of GLAT in Appendix~\ref{sec:appendix_glat}.}: First, its sampling rate cannot be decreased to zero during training, which exists the issue of \textit{exposure bias}.
Second, it still heavily relies on a teacher model for further improvements~\cite{glat}.




\paragraph{Latent Transformer.}\label{ss:latent}
To alleviate the multi-modality problem, \citet{lt,lv_nar,flowseq,cnat} propose Latent Transformer (LT), introducing latent variables $\z$ for NAT predictions as:
\begin{equation}\label{eqn:lt}
     p_\text{LT}(\y|\x) = \int_{\z} p(\z|\x)\cdot p(\y|\z,\x).
\end{equation}
where $p_\text{LT}(\y|\x)$ is always trained by variational inference~\cite{flowseq} or discretization techniques~\cite{lt}.
Such latent variables are decomposed from the target sentence, which is informative to determine the mode of the sentence and alleviates the multi-modality problems.


Although Latent Transformer models improve performance in terms of BLEU score, their used autoregressive predictor~\cite{lt,cnat} or deep iterative transformation~\cite{lv_nar,flowseq} for predicting latent variables unavoidable sacrifice the overall decoding efficiency.
Besides, they do not explicitly build the interdependencies among the outputs.

\section{Proposed Method: \method}
In this section, we present \method. 
\method follows Latent Transformer models~\cite{lt,cnat} but introduces glancing training~\cite{glat} with the discrete latent variables.
Our intuitions are as follows:

First, compared to the words, the introduced discrete latent variables may have fewer modes than words and be informative to determine the modes of the sentences.
In such a case, we can directly learn the discrete latent variables by the Glancing Transformer~\cite{glat}, keeping competitive inference efficiency.
More importantly, we can employ the latent variables to invoke glancing training for modeling the target sentences, which is informative enough to reduce the multi-modality problem of original sentences.
Besides, glancing at latent variables also works robustly due we can obtain the latent variables during inference.




\subsection{Introducing Discrete Latent Variables for Modeling Target Categorical Information}
In this part, we state the structure of \method, which introduces a small set of discrete latent variables for a NAT model, basically following \citet{lt,vqvae,cnat}. 

Let $K$ be the size of the discrete latent space and let $[K]$ denote the set $\{1,2,\cdots,K\}$. 
For each target sentence $\y=(y_1,y_2,\cdots,y_m)$, we use a same-length latent variable sequence for modeling it as:
\begin{equation}\label{eqn:mlt}
    p(\y|\x) = \sum_{\z} p_\theta(\z|\x)\cdot \prod_{t=1}^{m}p_\theta(y_{t}|\z,\x),
\end{equation}
where $\z=(z_1,z_2,\cdots,z_m)$ and $z_i \in [K]$, $\theta$ is the model parameters.

\paragraph{Discretization.}
For discretizing target sentences to latent variables, we use \textit{vector quantization}~\cite{vqvae}, which works by dividing a large set of origin vector representations into small groups.
We assign each token $y_i$ with a group $j\in[K]$ that has the nearest distance to its representation:
\begin{equation}
\label{eqn:vq}
z_i  =\argmin_{j\in [K]} || \operatorname{repr}(y_i) - \Q_j ||_{2},
\end{equation}
where $\Q\in\mathbb{R}^{\text{K}\times d_\text{model}}$ is the maintained representations and $d_\text{model}$ is its dimension. 
We use the embedding as $\operatorname{repr}(y_i)$, refer to~\citet{cnat}.
Finally, the model is trained to minimize
\begin{equation}\label{eqn:lt_loss}
    \mathcal{L}_\text{LT}=\mathcal{L}_\text{LP} + \mathcal{L}_\text{WP},
\end{equation}
where $\mathcal{L}_\text{WP}$ and $\mathcal{L}_\text{LP}$ are the prediction loss for words $\y$ and latent variables $\z$, respectively.

The maintained representations $\Q$ are updated with an exponential moving average over a mini-batch of target tokens $\{y_1,\cdots,y_i,\cdots\}$:
\begin{equation}\label{eqn:ema}
    \begin{split}
         c_j &\gets \lambda c_j + (1-\lambda) \sum_{i} \mathbbm{1}[z_i=j],\\
        \Q_j &\gets \lambda \Q_j + (1-\lambda)\sum_{i}\frac{\mathbbm{1}[z_i=j]\operatorname{repr}(y_i)}{c_j}
    \end{split}
\end{equation}
where $c_j$ is assigned count for group $j$, and we set decay parameter $\lambda=0.999$ in our experiments.

\begin{figure}[t]
\centering
\includegraphics[width=0.95\linewidth]{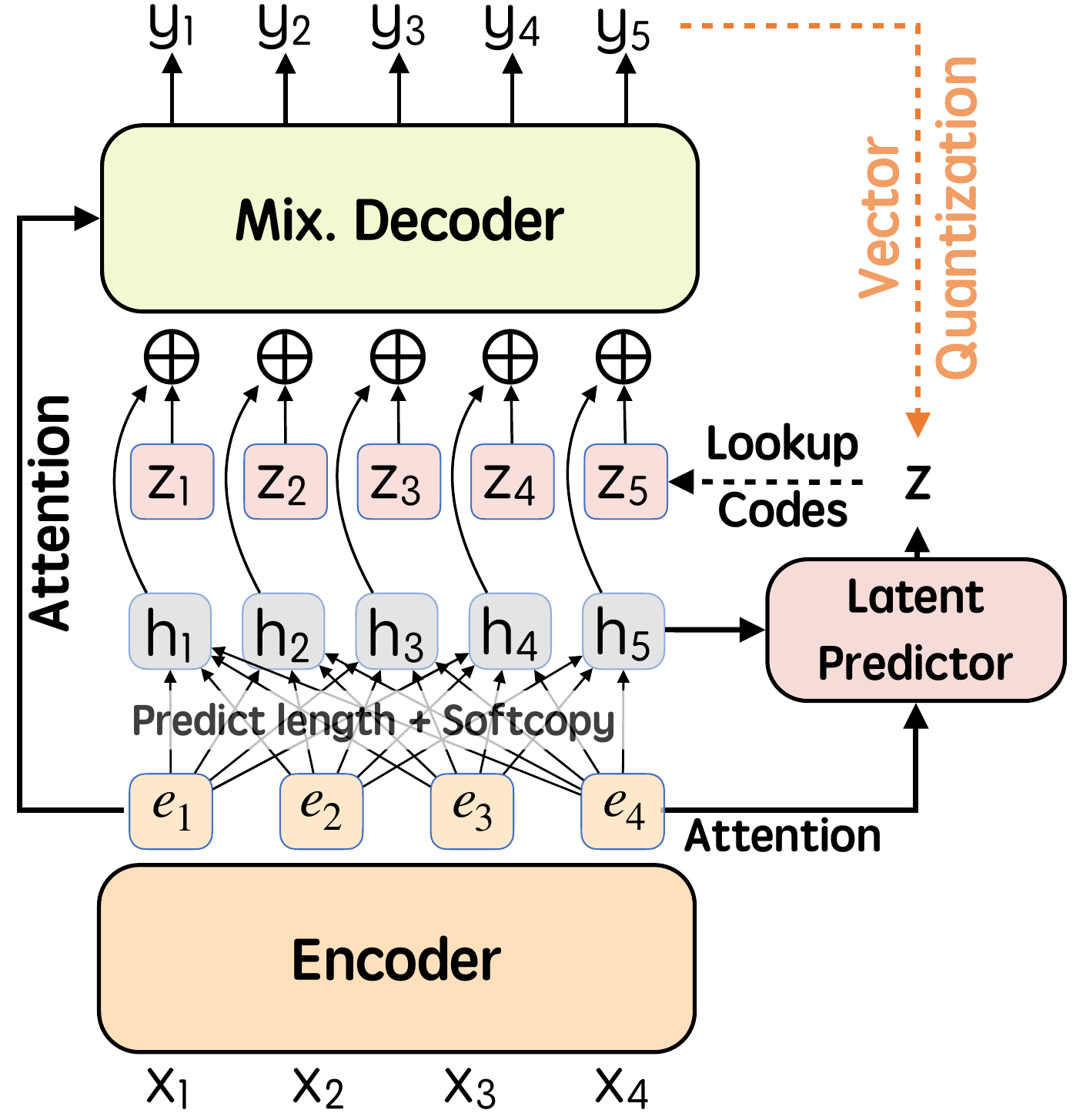}
\caption{\textbf{Model architecture of \method.} $\bigoplus$: Position-wise mix $h_i$ and  representation of $z_i$ by a gated neural network.}
\label{fig:arch}
\end{figure}

\begin{figure*}[t]
    \centering
    \begin{subfigure}[b]{0.375\linewidth}
    \centering
    \includegraphics[width=1.0\linewidth]{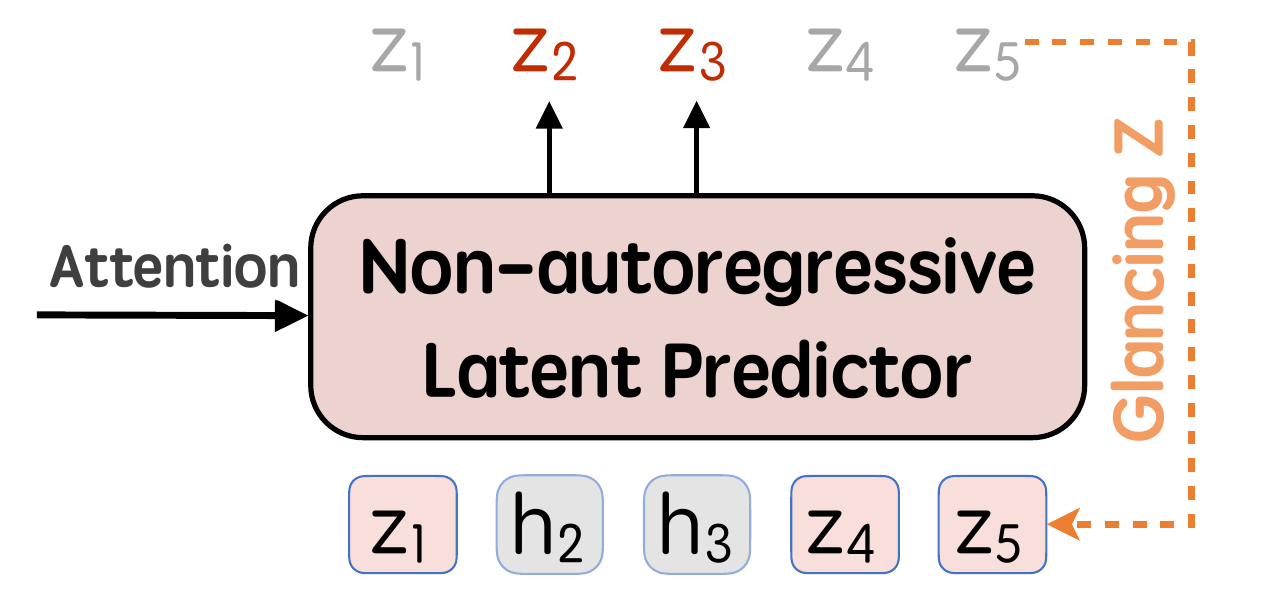}
    \caption{$\mathcal{L}^\text{GLT}_\text{LP}$ for Latent Predictor}
    \label{fig:lp}
    \end{subfigure}
    \hspace{.1\linewidth}
    \begin{subfigure}[b]{0.375\linewidth}
    \centering
    \includegraphics[width=1.0\linewidth]{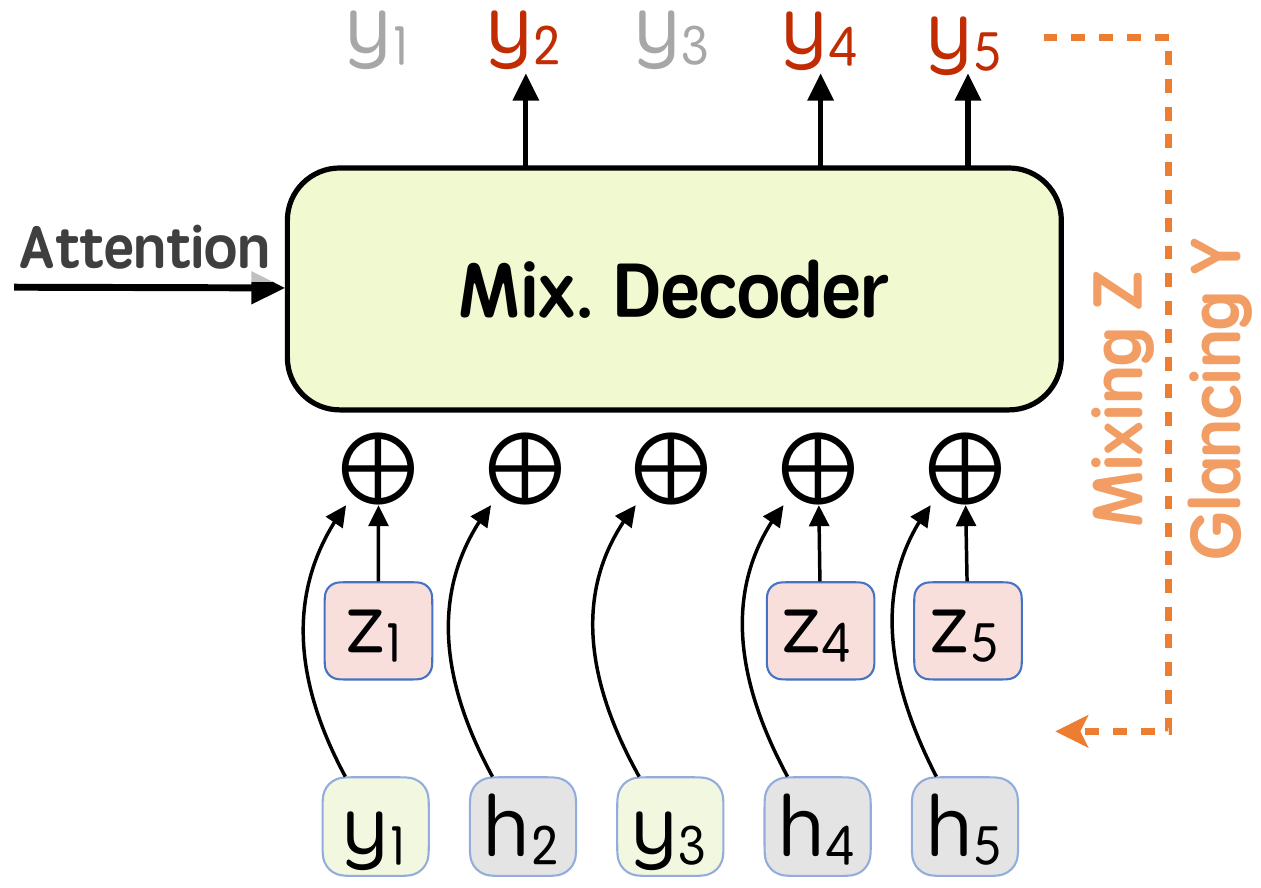}
    \caption{$\mathcal{L}^\text{GLT}_\text{WP}$ for Mixture Decoder}
    \label{fig:mix}
    \end{subfigure}
\caption{\textbf{Training the latent predictor and mixture decoder by glancing at discrete latent variables.}}
\end{figure*}

\paragraph{Architecture.}
As shown in Figure~\ref{fig:arch}, \method mainly consists of an encoder $\F_\text{ENC}$~(NAT Encoder), a latent predictor $\F_\text{LP}$~(NAT Predictor), and a decoder $\F_\text{DEC}$~(Mix. Decoder). 
We parameterize them with the multi-head attention-based encoder or decoder, similar to Transformer~\cite{transformer}.
Their functions can be formalized as:
\begin{equation*}
    \begin{split}
            (e_1,e_2,\cdots,e_n)&\gets \F_\text{ENC}(x_1,x_2,\cdots,x_n),\\
            (h_1,h_2,\cdots,h_m)&\gets \operatorname{softcopy}(e_{1:n}),\\
                 p_\theta(\z|\x)&\gets \F_\text{LP}(h_{1:m},e_{1:n}),\\
              p_\theta(\y|\z,\x)&\gets \F_\text{DEC}(z_{1:m},h_{1:m},e_{1:n}),
    \end{split}
\end{equation*}
where we use an extra module $\F_\text{LEN}$ to predict the target length $m$ and initialize the decoder inputs $H=(h_1,h_2,\cdots,h_m)$ with the $\operatorname{softcopy}$~\cite{imitate_nat} mechanism.



\subsection{Glancing at Discrete Latent Variables for Parallel Sequence Decoding}\label{ss:mixture}
The small number~($K<128$) of discrete latent variables can capture high-level categorical information of the target words, supporting better learning design for parallel sequence decoding.

Our first insight is that we can learn to non-autoregressively predict the discretized latent variables directly without the help of distillation.
Specifically, we parameterize the $\F_\text{LP}$ in a non-autoregressive fashion and use a glancing training technique~\cite[GLT,][]{glat} for optimizing it, as shown in Figure~\ref{fig:lp}:
\begin{equation}\label{eqn:lp_loss}
    \begin{split}
    \mathcal{L}^\text{GLT}_\text{LP} &= -\log p_\theta(\uno{\z}|\obs{\z},\x)\\
    \end{split}
\end{equation}
where $\obs{\z}$ is uniformly sampled from $\z$, refer to \citet{glat}.
We provide more training details of \method in Appendix~\ref{s:appendix_structures}.

Our next insight is modeling the sentence based on the sampled latent variables $\obs{\z}$ rather than $\z$, namely, glancing at $\obs{\z}$ for optimizing $\F_\text{DEC}$:
\begin{equation}\label{eqn:glz_loss}
    \mathcal{L}_\text{WP} =-\log  p_\theta(\y|\obs{\z},\x). \\
\end{equation}
We find Eqn.~(\ref{eqn:glz_loss}) works robustly in experiments and analyze it in Section~($\S$~\ref{ss:analysis}).

As shown in Figure~\ref{fig:mix}, we eventually employ words to invoke glancing training for minimizing $\mathcal{L}_\text{WP}$, namely we optimize the
$\F_\text{DEC}$ by minimizing\begin{equation}\label{eqn:mix_loss}
    \mathcal{L}^\text{GLT}_\text{WP} = -\log  p_\theta(\overline{\obs{\y}}|\obs{\z},\obs{\y},\x),
\end{equation}
where $\obs{\y}$ and $\obs{\z}$ are the sampled target tokens and discrete latent variables.

\paragraph{Overall Training Loss.} 
Our full-fledged loss includes latent variable prediction, sentence reconstruction, and length prediction losses:
\begin{equation}
    \mathcal{L} = \mathcal{L}^\text{GLT}_\text{WP} + \mathcal{L}^\text{GLT}_\text{LP} + \alpha \mathcal{L}_\text{LEN},
    \label{eqn:loss}
\end{equation}
where $\alpha=0.1$ are the hyperparameters to adjust the importance of length prediction loss $\mathcal{L}_\text{LEN}$.

\subsection{Inference}
In inference phase, \method predicts the target length, latent variables, and sentence in turn.

For the target length, \method first predicts the target length $m$ with the length predictor $\F_\text{LEN}$.
To avoid the length prediction errors during inference, \method expands the length $m$ to a ranges~(we use $[m-3,\cdots,m+2]$, total six candidates in our experiments).

Then, \method predicts the latent variables $\hat{\z}$ with $\argmax_{\z} p_\theta(\z|\x)$ and sentence $\hat{\y}$ with $\argmax_{\y} p_\theta(\y|\hat{\z},\x)$ for each candidate.

Similar to \citet{flowseq}, \method also ranks the candidates by itself~(\textit{self-reranking}) and chooses the highest score output with: 
\begin{equation}
    \hat{\y} =\argmax_{\y} p_\theta(\y|\hat{\z},\x)\cdot \gamma^{|\y|}
\end{equation}
where $\gamma$ is the length penalty ratio to avoid the length bias, and $|\y|$ denotes the length of $\y$.
%

\section{Experiments}\label{s:exp}
We conduct experiments on several generation tasks, including machine translation, paraphrase generation, and dialog generation.

\subsection{Experimental Setup}\label{ss:setup}
\paragraph{Dataset.}We chose the most popular benchmarks for each task: 
\begin{compactitem}
    \item \textbf{Machine Translation~(MT)}: 
    We follow previous practices in NAT models and use the WMT14 English~(EN) $\leftrightarrow$ German~(DE) corpus~(4.5M sentence pairs) and the IWSLT14 German~(DE) $\to$ English~(EN) corpus~(160K sentence pairs) to validate our proposed model.
    We obtain the datasets following the instruction open-sourced in \texttt{fairseq}\footnote{\url{https://github.com/pytorch/fairseq}}. 
    In detail, we first tokenize the datasets with \texttt{Moses} script.
    Then, we use 37,000 and 10,000 operations to split the words into byte-pair encodings~\citep[BPE,][]{bpe} in WMT14 and IWSLT14 datasets, respectively.
    We also share subword embeddings between the source and target language for each dataset.
    \item \textbf{Paraphrase Generation~(PG)}: 
    We use the Quora\footnote{\url{https://www.kaggle.com/c/quora-question-pairs/data}} dataset to evaluate the paraphrase generation task.
    The Quora dataset contains around 135K labeled paraphrases pairs. Following the standard dataset split, we sample 100K sentence pairs from the labeled paraphrases as training data and hold out 30K pairs for testing, the remaining about 5K pairs for validation. 
    Like the MT tasks, we tokenize the corpus with Moses scripts and split the words into BPE units with total 32K operations.
    \item \textbf{Dialog Generation~(DG)}: We conduct the dialog generation experiments on the DailyDialog dataset~\cite{dialog}. 
    We obtain the processed DailyDialog dataset from \citet{bao2020plato}\footnote{\url{https://github.com/gmftbyGMFTBY/MultiTurnDialogZoo}}. 
    The training set contains 87,170 sentence pairs (11,118 dialogues). 
    The validation and testing set in the dataset contain 8069 pairs (1000 dialogues) and 7740 pairs (1000 dialogues), respectively.
\end{compactitem}
Note that these tasks emphasize different aspects.
The task of MT aims to transfer bilingual sentences with semantically invariant conditions.
The PG task differs from machine translation and works on mode transformation in the same language, whose goal is to synthesize a sentence different from the original input but conveys the same meaning.
The DG task is most challenging due to the complex generation goal.

\paragraph{Implementations.} 
We compare \method with Transformer~\citep{transformer}, NAT~\citep{nat}, and GLAT~\cite{glat} models. 
We implement them based on the open-source framework \texttt{fairseq}~\citep{fairseq}.

For machine translation tasks, we use the base setting ($d_\text{model}=512$, $d_\text{hidden}=2048$, $\text{dropout}=0.1$, $n_\text{head}=8$, and $n_\text{layer}=6$) of Transformer~\cite{transformer} for WMT14 dataset and a smaller setting ($d_\text{model}=512$, $d_\text{hidden}=1024$, $\text{dropout}=0.3$, $n_\text{head}=4$, and $n_\text{layer}=6$) for IWSLT14 dataset. 
The number of layers in \method decoder and latent predictor are both set to 4 in experiments.
We use inverse square root learning rate scheduling for WMT14 and a linear annealing learning rate from $3.0\times10^{-4}$ to $1.0\times10^{-5}$ in 250K steps for IWSLT14.
The models are optimized with Adam~\citep{adam} optimizer ($\beta_1=0.9, \beta_2=0.999$) in 300K steps for WMT14 and 250K steps for IWSLT14. 
As for the ratio $\tau$ that used in glancing sampling, we linear anneal the ratio from $0.5$ to $0.3$ in whole training steps.
The mini-batch in each step consists of 2K tokens for IWSLT14 and 64K tokens for WMT14.

Since the scale of the Quora and DailyDialog datasets are close to the IWSLT14, we keep the same setting to the IWSLT14, such as the Adam, learning rate~(linear annealing from $3.0\times10^{-4}$ to $1.0\times10^{-5}$), and batch size~(2K tokens).

\paragraph{Evaluation.}
To validate the effectiveness of our proposed method, we evaluate it in terms of quality and efficiency.
We use tokenized and cased BLEU scores~\cite{bleu}\footnote{We evaluate BLEU using \texttt{fairseq\_score} script.} to evaluate the generation quality of MT and PG tasks.
For dialog generation, we also include BLEU-1 and BLEU-2 scores for analysis.
Following the common practices~\cite{nat,glat}, we measure the decoding latency of each model by decoding sentence by sentence and compute the speedup compared with the autoregressive Transformer~(AT) model to reflect its decoding efficiency.
We highlight the \textbf{best NAT} result.

\begin{table*}[t]
\centering
\small
\tabcolsep 4.5pt
\begin{tabular}{lcccccccrr}
\toprule
\multirow{2}{*}{\textbf{Models}} & \multicolumn{2}{c}{\textbf{WMT14}}      & \textbf{IWSLT14}       & \multirow{2}{*}{\textbf{Quora}} & \multicolumn{3}{c}{\textbf{DailyDialog}} & \multirow{2}{*}{\textbf{Latency}$^\downarrow$} & \multirow{2}{*}{\textbf{Speedups}$^\uparrow$}\\ \cmidrule{6-8}\cmidrule{2-4}
                       & \textbf{EN$\to$DE}      & \textbf{DE$\to$EN}     & \textbf{DE$\to$EN}     &  &  \textbf{BLEU-1}   &\textbf{BLEU-2}    & \textbf{BLEU} & \\    
\midrule
Transformer (AT)            & 27.17          & 31.53         & 34.29         & 27.97 &   31.40   & 10.70     & 5.05  &512.3 ms& 1.00 $\times$\\ 
\midrule
NAT                    & 10.78          & 15.19         & 17.77         & 24.65 &   \textbf{41.50}   & 1.40      & 0.01  & \textbf{33.5} ms& \textbf{15.29} $\times$ \\
GLAT                   & 16.71          & 24.78         & 29.07         & 27.01 &   39.50   & 26.20     & 26.13 & \textbf{33.5} ms& \textbf{15.29} $\times$ \\
\method                &\textbf{24.71}  &\textbf{29.16} &\textbf{32.31} &\textbf{29.11} & 41.00 &\textbf{28.30}  &\textbf{27.50} & 45.3 ms& 11.31 $\times$\\
\bottomrule
\end{tabular}
\caption{\textbf{Main results of different models on the test set of each dataset.} We measure the decoding latency and speedups on the WMT14 EN$\to$DE test set.}
\label{tab:main}
\end{table*}

\subsection{Main Results}
We can see from Table~\ref{tab:main} that our~\method almost outperforms all the NAT baselines (NAT and GLAT) in generation quality on all tasks while keeping a competitive decoding speedup to the autoregressive counterpart.

\paragraph{Machine Translation.}
As seen, without the help of an AT model for training, the vanilla NAT and advanced GLAT model only obtain inferior generation quality.
In contrast, \method achieves competitive generation quality in machine translation tasks, indicating that the introduced latent variables effectively reduce the multi-modality issue and support glancing training well.
It narrows the performance gap between non-autoregressive decoding and autoregressive decoding from 11.46~(GLAT vs. AT) to 2.34~(\method vs. AT) BLEU points on WMT14 EN$\to$DE task while keeping a high-speed decoding efficiency.

\paragraph{Paraphrasing.}
Unlike the translation task, the performance gap between non-autoregressive and autoregressive decoding on the paraphrase generation task is minor~(NAT vs. AT, $-3.32$ BLEU points, GLAT vs. AT, $-0.96$ BLEU points ). 
Nevertheless, introducing discrete latent variables still is helpful to obtain a better performance.
\method realizes a non-autoregressive model with better performance than the autoregressive model on Quora~(\method vs. AT, $+1.14$ points).

\paragraph{Dialog Generation.}
We can see a different trend on the DailyDialog dataset --- an AT model performs poorly than NAT models.
Both GLAT and \method outperform the AT model in BLEU-1, BLEU-2, and BLEU scores, indicating that these models recall more reference tokens and organize the tokens well.

We conjecture that the weak and indirect association between the inputs and outputs of the dialogue results in this unusual phenomenon.
Specifically, the weak connection may encourage the AT model to predict the tokens by paying more attention to their history outputs, which degenerate to a target-side language model.
In contrast, the NAT models do not have this fast track, pushing them to pay more attention to the inputs and recall more target tokens.
We further find that there are so-called \textit{safe response}~\cite{li2016diversity} in AT's outputs, which verify our conjecture.

\begin{table}[t]
\centering
\small
\tabcolsep 2.5pt
\begin{tabular}{lrccr}
\toprule
\multirow{2}{*}{\textbf{Models}} & \multicolumn{2}{c}{\textbf{WMT14}} & \textbf{IWSLT14}& \multirow{2}{*}{\textbf{Speedups$^\uparrow$}} \\ \cmidrule{2-4}
                      & \textbf{EN$\to$DE} & \textbf{DE$\to$EN} & \textbf{DE$\to$EN}    \\
\midrule
CMLM$_1$    & $^*$10.88       & -       & -     & -                          \\ 
CMLM$_4$    & $^*$22.06       & -       & -     & $^\dag$9.79~$\times$         \\ 
CMLM$_{10}$ & $^*$24.65       & -       & -     & $^\dag$3.77~$\times$        \\ 
LevT$_{2.05}$ & 24.43       & -       & -     & 2.93~$\times$        \\ 
\midrule
LV-NAR   & 11.80       & -       & -     & \textbf{22.30}~$\times$         \\
SynST    & 20.74       & 25.50   & 23.82 & 4.86~$\times$            \\
Flowseq & 20.85       & 25.40   & -     & $^\ddag$1.10~$\times$         \\
CNAT       & 21.30       & 25.73   & 29.81 & 10.37~$\times$         \\
\midrule
AT                      & 27.17       & 31.53   & 34.29 & 1.00~$\times$    \\ 
NAT                     & 10.78       & 15.19   & 17.77 & 15.29~$\times$   \\
GLAT                    & 16.71       & 24.78   & 29.07 & 15.29~$\times$     \\
\method                &\textbf{24.71} &\textbf{29.16} & \textbf{32.31}  & 11.31~$\times$  \\
\bottomrule
\end{tabular}
\caption{\textbf{BLEU scores and speedups of different models trained with raw datasets on machine translation tasks.} We quote some results from $^*$\citet{flowseq}, $^\dag$\citet{jm_nat}, $^\ddag$\citet{glat}, and the original paper. CMLM$_n$ and LevT$_n$: using $n$ iterations during inference.  $-$: no corresponding results.}
\label{tab:translation}
\end{table}

\paragraph{More Comparisons.}
we further compare the advanced NAT models that builds upon latent variables or iterative refinement in machine translation tasks:
\begin{compactitem}
    \item NATs w/ latent variables: LV-NAR~\cite{lv_nar}, SynST~\cite{syn_st}, Flowseq~\cite{flowseq}, and CNAT~\cite{cnat}.
    \item Iterative NATs: CMLM~\cite{cmlm} and LevT~\cite{levT}.
\end{compactitem}



Table~\ref{tab:translation} shows that introducing latent variables (LV-NAR, Flowseq, and CNAT) or decoding with multiple iterations (CMLM and LevT) both improve non-autoregressive decoding in translation quality.
However, iterative refinements or deep transformations always sacrifice decoding efficiency.
In contrast, the proposed \method outperforms all NAT models with a relatively low cost, keeping a competitive speedup over autoregressive Transformer~(AT).
Specifically, \method with one-pass decoding narrows the performance gap to the AT from 5.87 BLEU points to 2.34 BLEU points on the WMT14 EN$\to$DE test set.

\begin{figure}[t]
\centering
\includegraphics[width=\linewidth]{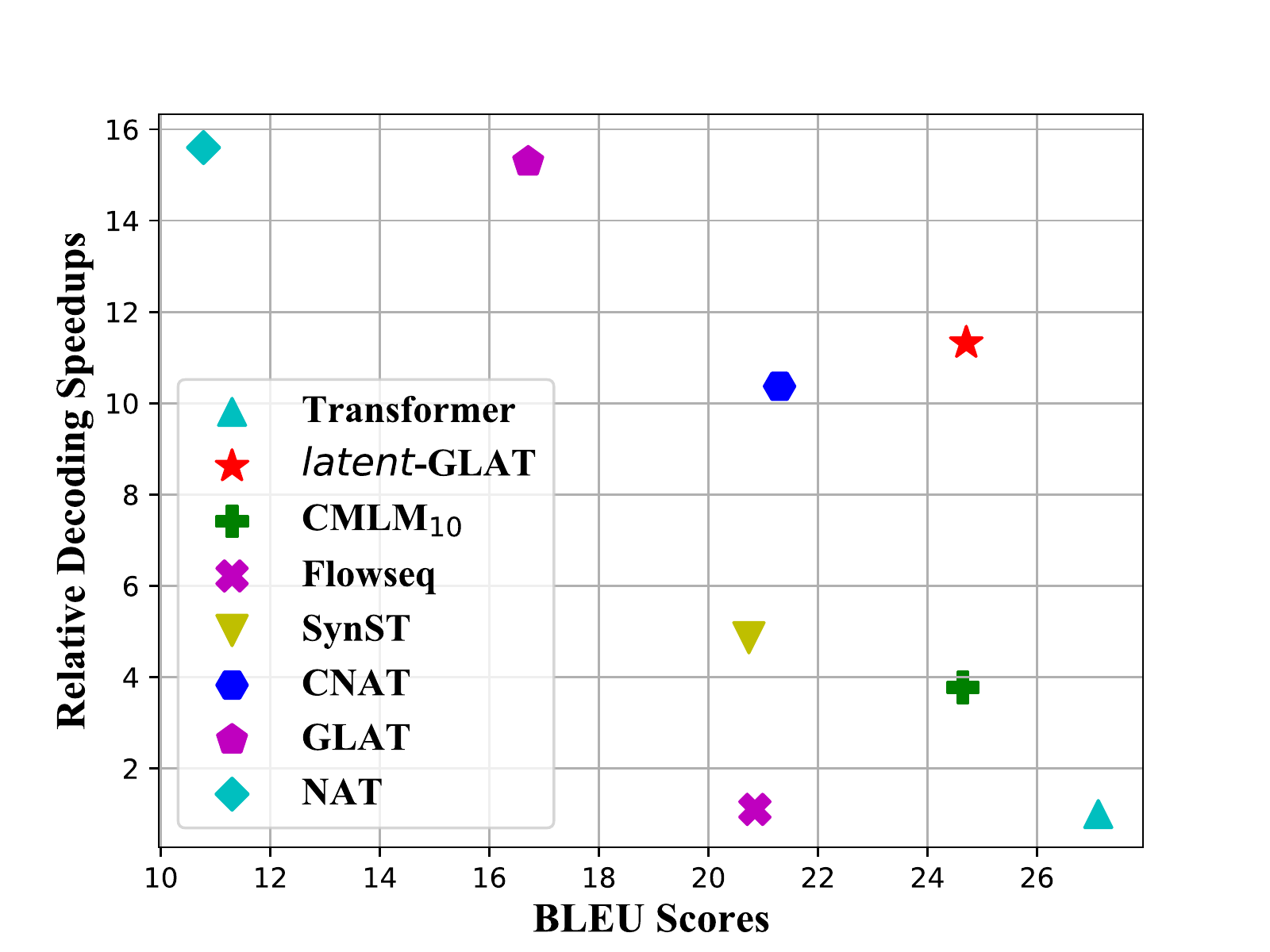}
\caption{\textbf{BLEU scores and their relative decoding speedups of different models on WMT14 EN$\to$DE test set.}
Note that we evaluate the speedups with a single GTX 1080-Ti GPU and include the results with the same evaluating hardware for fair comparisons.
}
\label{fig:speed}
\end{figure}

\paragraph{Decoding efficiency.}
We can see there is a trade-off between the translation quality and decoding efficiency in Table~\ref{tab:translation}. 
We thus present the scatter plot of different models in Figure~\ref{fig:speed}, showing the trend of translation quality and decoding efficiency.

As seen, \method is located on the top-right of the baselines. 
It outperforms the baselines in the BLEU score if decoding speedup is fixed and in decoding speedup if the BLEU score is fixed.

\subsection{Analysis}\label{ss:analysis}
We now turn to verify our intuition that \method can alleviate the multi-modality problem.

\begin{table}[t]
\centering
\small
\tabcolsep 3pt
\begin{tabular}{lcccc}
\toprule
\multirow{2}{*}{\textbf{Methods}}  & \multicolumn{2}{c}{\textbf{WMT14}}      & \multicolumn{1}{c}{\textbf{IWSLT14}}       &\multirow{2}{*}{\textbf{Avg $\Delta$}$^\downarrow$}\\ \cmidrule{2-4} 
                  & \multicolumn{1}{c}{\textbf{EN$\to$DE}}   & \multicolumn{1}{c}{\textbf{DE$\to$EN}}        & \multicolumn{1}{c}{\textbf{DE$\to$EN}}     &  \\
\midrule
 NAT                   & 10.78       & 15.19            & 17.77             &\multirow{2}{*}{+6.58} \\
 \quad w/ KD   & 17.69      & 22.02           &  23.78                                     \\
\cmidrule{1-5}
GLAT                    & 16.71       & 24.78           & 29.07                 &\multirow{2}{*}{+5.19} \\
\quad w/ KD        & 25.21 & 29.84  & 31.07    \\
\midrule
Flowseq                 & 20.85       & 25.40            & 24.75             &\multirow{2}{*}{+2.87} \\
\quad w/ KD       & 23.72       & 28.39            &  27.55                                      \\
\cmidrule{1-5}
CNAT                    & 21.30       & 25.73           & 29.81         &\multirow{2}{*}{+3.08}\\
 \quad w/ KD        & 25.56  & 29.36  & 31.15      \\ 
\cmidrule{1-5}
\method                 & 24.71          & 29.16    & 32.31                 &\multirow{2}{*}{\textbf{+0.95}}\\
 \quad w/ KD        & \textbf{26.64} & \textbf{29.93}   &\textbf{32.47}\\
\bottomrule
\end{tabular}
\caption{\textbf{BLEU scores of NAT models trained with (or without) knowledge distillation~(KD) on translation tasks.}}
\label{tab:translation_kd}
\end{table}

\paragraph{\method largely alleviates the sentence-level multi-modal problem.}
Previous researches~\cite{nat,flowseq,glat,cnat} always utilize a Transformer model as a teacher for training NAT models, namely sequence-level knowledge distillation~\cite{kim2016sequence}, which can directly reduces the sentence-level multi-modal phenomenon in datasets.
Therefore, we use the average gains from the knowledge distillation to reflect the ability of the NAT models to overcome this issue.

As seen in Table~\ref{tab:translation_kd}, the pure NAT models heavily rely on knowledge distillation.
By introducing the target information with the latent variables (Flowseq and CNAT) or sampled tokens (GLAT), the NAT models improve its' ability to overcome the multi-modality issue.
Our proposed \method well combines the above two techniques.
It obtains only 0.95 BLEU points average gains and validates our motivation.

\begin{table}[t]
\centering
\small
\tabcolsep 2.5pt
\begin{tabular}[c]{llcc}
\toprule
\textbf{Datasets} &\multicolumn{1}{c}{\textbf{Configuration}~($\bm d$)}&\textbf{$C_\text{TOK}(d)$}&\textbf{$C_\text{SEN}(d)$} \\
\midrule 
\multirow{3}{*}{WMT14}& Inputs $\leftrightarrow$ Raw outputs     & 2.19          & 3.03 \\
&Inputs $\leftrightarrow$ AT outputs   & 1.38          & 2.13\\
&Inputs $\leftrightarrow \z$      & \textbf{1.01} & \textbf{1.35}\\
\midrule
Quora  & Inputs $\leftrightarrow$ Raw outputs  & 0.86          & 1.48 \\
\midrule 
DailyDialog  & Inputs $\leftrightarrow$ Raw outputs  & 1.19          & 4.23 \\
\bottomrule
\end{tabular}
\caption{\textbf{Token-level or sentence-level complexity of different text generation datasets.} The higher $C_\text{TOK}(d)$ or $C_\text{SEN}(d)$, the more complex.}
\label{tab:complexity}
\end{table}

\paragraph{Discrete latent variables have fewer modes than raw sentences.}
To validate our intuition that the introduced latent variables are easier to predict than tokens, we refer to \citet{zhou2019understanding} to compute the complexity metrics on each dataset according to alignment relations.
Specifically, we use the \textit{fast\_align}\footnote{\url{https://github.com/clab/fast_align}} toolkit to align source input $\x$ and target outputs $\y$ or discretized latent variable sequences $\z$.
Then, we compute the token-level complexity $C_\text{TOK}(d)$ and the sentence-level complexity $C_\text{SEN}(d)$ according to  \citet{zhou2019understanding}.
These metrics can trivially understand as the number of valid candidates for each input.

As shown in Table~\ref{tab:complexity}, the latent variables have the lowest complexity in both token-level complexity and sentence-level complexity.
In other words, predicting the latent variable sequences is effortless than predicting others, which is consistent with our intuition.
Although we obtain a lower complexity dataset by filtering the datasets with an autoregressive model~(AT outputs versus Raw outputs), they may introduce model error and need extra training for AT model.
In contrast, the discrete latent variables are simple and informative enough to serve as a springboard for modeling target sentences.


\begin{table}[t]
\centering
\small
\tabcolsep 4pt
\begin{tabular}{ccccl}
\toprule
\multirow{2}{*}{\textbf{L\#}} & \multirow{2}{*}{\textbf{Introduce} $\z$} & \multicolumn{2}{c}{\textbf{Glancing Training}}& \multicolumn{1}{c}{\multirow{2}{*}{\textbf{BLEU ($\Delta$)$^\uparrow$}}} \\ \cmidrule{3-4}
                        &                       & \textbf{with} $\z$       & \textbf{with} $\y$  & \\
\midrule
1                      &                    &                   &                    & 12.60 \\
2                      & \hit               &                   &                    & 13.43~(+0.83)\\
3                      &                    &                   &  \hit              & 17.11~(+4.51)\\
4                      & \hit               &                   &  \hit              & 18.88~(+6.20)\\
5                      & \hit               &  \hit             &                    & 22.35~(+9.75)\\
6                      & \hit               &  \hit             &  \hit              & \textbf{23.64}~(+11.04)\\
\bottomrule
\end{tabular}
\caption{\textbf{BLEU scores of different \method configurations on the WMT14 EN$\to$DE valid set.}}
\label{tab:ablation}
\end{table}

\paragraph{Glancing with latent variables improves the performance with a large margin.}
We can see in Table \ref{tab:ablation} that introducing latent variables both obtain performance gains to their counterpart (L\#2 vs. L\#1, $+0.83$ points, and L\#4 vs. L\#3, $+1.69$ points).
As expected,  the gains are largely improved while adopting the glancing training with discrete latent variables~(L\#5 vs. L\#1, $+9.75$ points), which already outperforms glancing training with the reference token~(L\#5 vs. L\#4, $+3.55$ points).
Finally,  we jointly perform glancing training with the reference tokens and discrete latent variables, achieving the best result~(L\#6 vs. L\#1, $+11.04$ points).


\begin{table}[t]
\centering
\small
\tabcolsep 4pt
\begin{tabular}{rcccccc}
\toprule
$\bm K$ & 8 & 16 & 32 & 64 & 128 & 256 \\
\midrule
\textbf{BLEU} (\%) & 20.80 & 22.16 & 22.61 & \textbf{23.64} & 23.26 & 21.94\\
\textbf{ACC}$_{\z}$ (\%)  & 61.20 & 53.10 & 43.57 &  39.24  & 36.39 & 33.84\\
\bottomrule
\end{tabular}
\caption{\textbf{Performances of \method with different $K$ on the WMT14 EN$\to$DE valid set.} We compute the accuracy~(ACC$_{\z}$) of latent prediction by taking the discretized latent variables as reference.}
\label{tab:k}
\end{table}

\begin{figure}[t]
    \centering
    \includegraphics[width=0.95\linewidth]{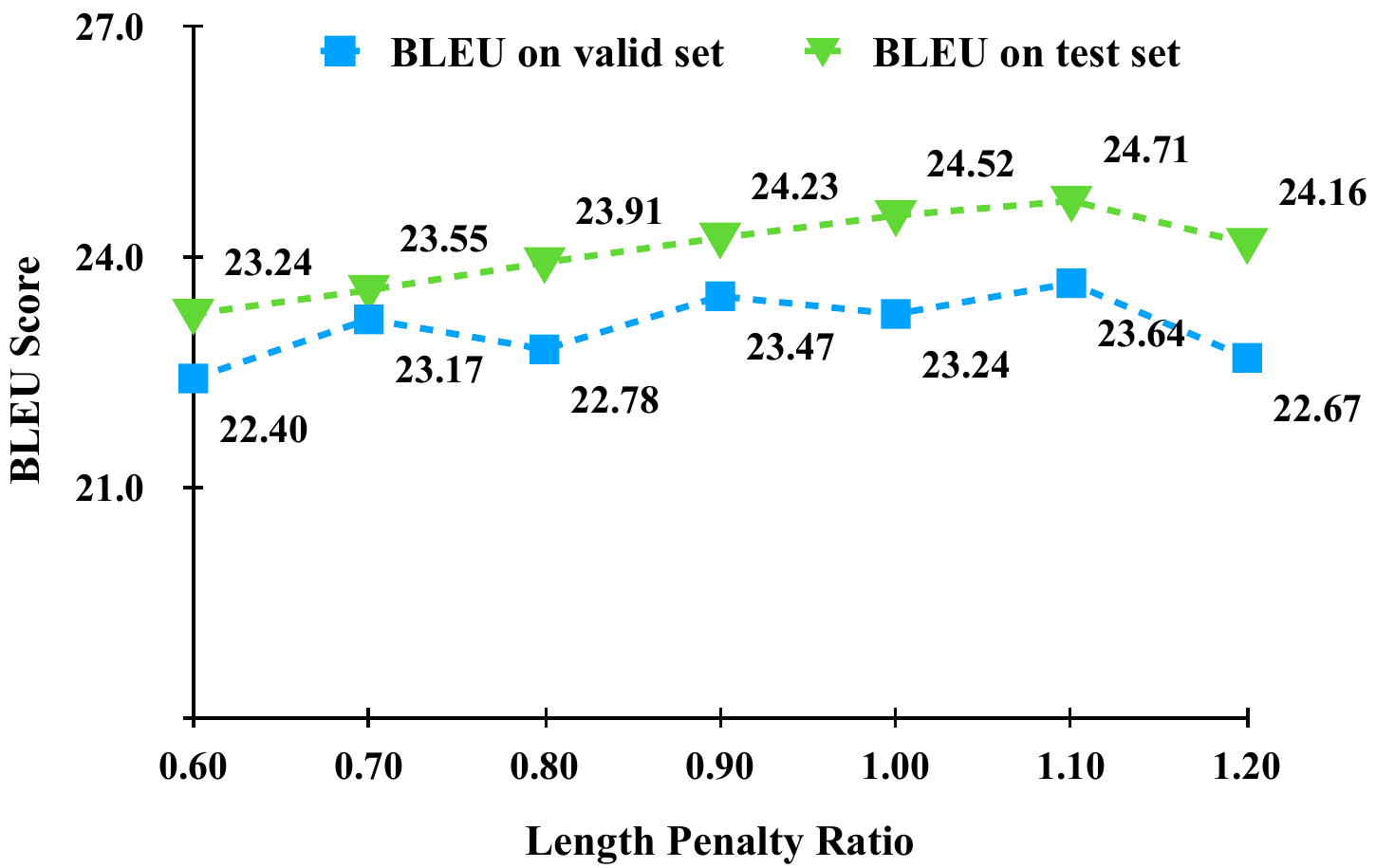}
    \caption{\textbf{BLEU scores of \method using different length penalty ratios on the WMT14 EN$\to$DE valid set.} We search the length penalty ratio $\gamma$ for \method while fixing the $K=64$.}
    \label{fig:ablation}
\end{figure}

\paragraph{Effects of $K$ and $\gamma$.} 
As shown in Figure~\ref{fig:ablation} and Table~\ref{tab:k}, we search the hyper-parameter of \method that the number of discrete latent variables and the length penalty ratio $\gamma$ according to the validation performance.
We notice that using more latent codes causes performance degradation during inference, in which the latent variables may degenerate to tokens and contains more prediction error during inference.
The \method implemented with 64 latent variables and $\gamma=1.1$ obtains the best result on WMT14 EN$\to$DE valid set.


\section{Related Work}
\citet{nat} first propose a non-autoregressive Transformer~(NAT) model for neural machine translation~(NMT) and begin to explore parallel decoding.
It abandons explicitly modeling word inter-dependencies to decode the tokens in parallel, significantly improving the inference speed.
However, its translation quality is inferior to the Transformer~\cite{transformer}.

To alleviate this performance degradation, many researchers work to enhance word dependency modeling, including imitation learning~\cite{imitate_nat,hint_nat}, curriculum learning~\cite{guo2020fine,liutask}, iterative refinements~\cite{iter_nat,cmlm,levT,jm_nat,huang2021non}, and a simplified autoregressive process~\cite{nat_crf}.
The most representative method is the glancing transformer model~\cite{glat}, which adaptively and progressively samples partial tokens as inputs and predicts the remaining tokens, effectively establishing the dependencies between the sampled tokens and the remaining tokens.
However, these models still rely on a teacher for training, which cannot directly learn the raw dataset that contains one-to-many multi-modality phenomenon.

Introducing latent variables~\cite{pnat,cnat} to organize the target sentence is also a helpful route. 
Among them, our method is close to \citet{lt,lv_nar,flowseq,syn_st,cnat}.
These methods decompose the latent variables~(hints) from the target sentence and divide the origin goal into two parts: modeling latent variables and modeling the target sentences based on latent variables.
It implicitly overcomes the multi-modality phenomenon of target sentences because the latent variables can largely determine the mode of the sentence.
However, these methods always model the latent variables with an autoregressive predictor, which naturally sacrifices the decoding efficiency.

Unlike them, our approach models the discrete latent variables in a non-autoregressive fashion and extends glancing training with the discrete latent variables.
As a result, \method accomplishes a competitive performance both in decoding efficiency and quality.

\section{Conclusion}
We propose \method, which can be directly trained without the help of knowledge distillation.
Specifically, we employ discrete latent variables to capture the word categorical information and divide the original goal into the latent variables modeling and word prediction tasks.
Then, we learn each task with the glancing training and encourage the model to build dependencies on the latent variables, which have fewer modes than the words and are also informative for modeling the target sentences.
Experiments results on machine translation, paraphrase generation, and dialogue generation tasks validate the effectiveness of our \method.

\section*{Acknowledgements}
We would like to thank the anonymous reviewers for their insightful comments. Shujian Huang is the corresponding author. This work is supported by National Science Foundation of China (No. U1836221, 6217020152).

\bibliography{custom}

\begin{thebibliography}{36}
\expandafter\ifx\csname natexlab\endcsname\relax\def\natexlab#1{#1}\fi

\bibitem[{Akoury et~al.(2019)Akoury, Krishna, and Iyyer}]{syn_st}
Nader Akoury, Kalpesh Krishna, and Mohit Iyyer. 2019.
\newblock \href {https://doi.org/10.18653/v1/P19-1122} {Syntactically
  supervised transformers for faster neural machine translation}.
\newblock In \emph{Proceedings of the 57th Annual Meeting of the Association
  for Computational Linguistics}, pages 1269--1281, Florence, Italy.
  Association for Computational Linguistics.

\bibitem[{Bahdanau et~al.(2015)Bahdanau, Cho, and Bengio}]{attn_seq_to_seq}
Dzmitry Bahdanau, Kyunghyun Cho, and Yoshua Bengio. 2015.
\newblock \href {http://arxiv.org/abs/1409.0473} {Neural machine translation by
  jointly learning to align and translate}.
\newblock In \emph{3rd International Conference on Learning Representations,
  {ICLR} 2015, San Diego, CA, USA, May 7-9, 2015, Conference Track
  Proceedings}.

\bibitem[{Bao et~al.(2020)Bao, He, Wang, Wu, and Wang}]{bao2020plato}
Siqi Bao, Huang He, Fan Wang, Hua Wu, and Haifeng Wang. 2020.
\newblock \href {https://doi.org/10.18653/v1/2020.acl-main.9} {{PLATO}:
  Pre-trained dialogue generation model with discrete latent variable}.
\newblock In \emph{Proceedings of the 58th Annual Meeting of the Association
  for Computational Linguistics}, pages 85--96, Online. Association for
  Computational Linguistics.

\bibitem[{Bao et~al.(2021)Bao, Huang, Xiao, Wang, Dai, and Chen}]{cnat}
Yu~Bao, Shujian Huang, Tong Xiao, Dongqi Wang, Xinyu Dai, and Jiajun Chen.
  2021.
\newblock \href {https://doi.org/10.18653/v1/2021.naacl-main.458}
  {Non-autoregressive translation by learning target categorical codes}.
\newblock In \emph{Proceedings of the 2021 Conference of the North American
  Chapter of the Association for Computational Linguistics: Human Language
  Technologies}, pages 5749--5759, Online. Association for Computational
  Linguistics.

\bibitem[{Bao et~al.(2019)Bao, Zhou, Feng, Wang, Huang, Chen, and Li}]{pnat}
Yu~Bao, Hao Zhou, Jiangtao Feng, Mingxuan Wang, Shujian Huang, Jiajun Chen, and
  Lei Li. 2019.
\newblock \href {https://arxiv.org/pdf/1911.10677.pdf} {Non-autoregressive
  transformer by position learning}.
\newblock \emph{arXiv preprint arXiv:1911.10677}.

\bibitem[{Chen et~al.(2019)Chen, Watanabe, Villalba, and
  Dehak}]{chen2019listen}
Nanxin Chen, Shinji Watanabe, Jes{\'u}s Villalba, and Najim Dehak. 2019.
\newblock \href {https://arxiv.org/pdf/1911.04908.pdf} {Listen and fill in the
  missing letters: Non-autoregressive transformer for speech recognition}.
\newblock \emph{arXiv preprint arXiv:1911.04908}.

\bibitem[{Gehring et~al.(2017)Gehring, Auli, Grangier, Yarats, and
  Dauphin}]{cnn_seq}
Jonas Gehring, Michael Auli, David Grangier, Denis Yarats, and Yann~N. Dauphin.
  2017.
\newblock \href {http://proceedings.mlr.press/v70/gehring17a.html}
  {Convolutional sequence to sequence learning}.
\newblock In \emph{Proceedings of the 34th International Conference on Machine
  Learning, {ICML} 2017, Sydney, NSW, Australia, 6-11 August 2017}, volume~70
  of \emph{Proceedings of Machine Learning Research}, pages 1243--1252. {PMLR}.

\bibitem[{Ghazvininejad et~al.(2019)Ghazvininejad, Levy, Liu, and
  Zettlemoyer}]{cmlm}
Marjan Ghazvininejad, Omer Levy, Yinhan Liu, and Luke Zettlemoyer. 2019.
\newblock \href {https://doi.org/10.18653/v1/D19-1633} {Mask-predict: Parallel
  decoding of conditional masked language models}.
\newblock In \emph{Proceedings of the 2019 Conference on Empirical Methods in
  Natural Language Processing and the 9th International Joint Conference on
  Natural Language Processing (EMNLP-IJCNLP)}, pages 6112--6121, Hong Kong,
  China. Association for Computational Linguistics.

\bibitem[{Gu et~al.(2018)Gu, Bradbury, Xiong, Li, and Socher}]{nat}
Jiatao Gu, James Bradbury, Caiming Xiong, Victor O.~K. Li, and Richard Socher.
  2018.
\newblock \href {https://openreview.net/forum?id=B1l8BtlCb} {Non-autoregressive
  neural machine translation}.
\newblock In \emph{6th International Conference on Learning Representations,
  {ICLR} 2018, Vancouver, BC, Canada, April 30 - May 3, 2018, Conference Track
  Proceedings}. OpenReview.net.

\bibitem[{Gu et~al.(2019)Gu, Wang, and Zhao}]{levT}
Jiatao Gu, Changhan Wang, and Junbo Zhao. 2019.
\newblock \href
  {https://proceedings.neurips.cc/paper/2019/hash/675f9820626f5bc0afb47b57890b466e-Abstract.html}
  {Levenshtein transformer}.
\newblock In \emph{Advances in Neural Information Processing Systems 32: Annual
  Conference on Neural Information Processing Systems 2019, NeurIPS 2019,
  December 8-14, 2019, Vancouver, BC, Canada}, pages 11179--11189.

\bibitem[{Guo et~al.(2020{\natexlab{a}})Guo, Tan, Xu, Qin, Chen, and
  Liu}]{guo2020fine}
Junliang Guo, Xu~Tan, Linli Xu, Tao Qin, Enhong Chen, and Tie-Yan Liu.
  2020{\natexlab{a}}.
\newblock Fine-tuning by curriculum learning for non-autoregressive neural
  machine translation.
\newblock In \emph{Proceedings of the AAAI Conference on Artificial
  Intelligence}, volume~34, pages 7839--7846.

\bibitem[{Guo et~al.(2020{\natexlab{b}})Guo, Xu, and Chen}]{jm_nat}
Junliang Guo, Linli Xu, and Enhong Chen. 2020{\natexlab{b}}.
\newblock \href {https://doi.org/10.18653/v1/2020.acl-main.36} {Jointly masked
  sequence-to-sequence model for non-autoregressive neural machine
  translation}.
\newblock In \emph{Proceedings of the 58th Annual Meeting of the Association
  for Computational Linguistics}, pages 376--385, Online. Association for
  Computational Linguistics.

\bibitem[{Hamming(1950)}]{hamming1950error}
Richard~W Hamming. 1950.
\newblock \href {https://ieeexplore.ieee.org/abstract/document/6772729} {Error
  detecting and error correcting codes}.
\newblock \emph{The Bell system technical journal}, 29(2):147--160.

\bibitem[{Huang et~al.(2022)Huang, Zhou, Za{\"\i}ane, Mou, and
  Li}]{huang2021non}
Chenyang Huang, Hao Zhou, Osmar~R Za{\"\i}ane, Lili Mou, and Lei Li. 2022.
\newblock \href {https://arxiv.org/abs/2110.07515} {Non-autoregressive
  translation with layer-wise prediction and deep supervision}.
\newblock In \emph{AAAI}.

\bibitem[{Kaiser et~al.(2018)Kaiser, Bengio, Roy, Vaswani, Parmar, Uszkoreit,
  and Shazeer}]{lt}
Lukasz Kaiser, Samy Bengio, Aurko Roy, Ashish Vaswani, Niki Parmar, Jakob
  Uszkoreit, and Noam Shazeer. 2018.
\newblock \href {http://proceedings.mlr.press/v80/kaiser18a.html} {Fast
  decoding in sequence models using discrete latent variables}.
\newblock In \emph{Proceedings of the 35th International Conference on Machine
  Learning, {ICML} 2018, Stockholmsm{\"{a}}ssan, Stockholm, Sweden, July 10-15,
  2018}, volume~80 of \emph{Proceedings of Machine Learning Research}, pages
  2395--2404. {PMLR}.

\bibitem[{Kim and Rush(2016)}]{kim2016sequence}
Yoon Kim and Alexander~M. Rush. 2016.
\newblock \href {https://doi.org/10.18653/v1/D16-1139} {Sequence-level
  knowledge distillation}.
\newblock In \emph{Proceedings of the 2016 Conference on Empirical Methods in
  Natural Language Processing}, pages 1317--1327, Austin, Texas. Association
  for Computational Linguistics.

\bibitem[{Kingma and Ba(2015)}]{adam}
Diederik~P. Kingma and Jimmy Ba. 2015.
\newblock \href {http://arxiv.org/abs/1412.6980} {Adam: {A} method for
  stochastic optimization}.
\newblock In \emph{3rd International Conference on Learning Representations,
  {ICLR} 2015, San Diego, CA, USA, May 7-9, 2015, Conference Track
  Proceedings}.

\bibitem[{Lee et~al.(2018)Lee, Mansimov, and Cho}]{iter_nat}
Jason Lee, Elman Mansimov, and Kyunghyun Cho. 2018.
\newblock \href {https://doi.org/10.18653/v1/D18-1149} {Deterministic
  non-autoregressive neural sequence modeling by iterative refinement}.
\newblock In \emph{Proceedings of the 2018 Conference on Empirical Methods in
  Natural Language Processing}, pages 1173--1182, Brussels, Belgium.
  Association for Computational Linguistics.

\bibitem[{Li et~al.(2016)Li, Galley, Brockett, Gao, and
  Dolan}]{li2016diversity}
Jiwei Li, Michel Galley, Chris Brockett, Jianfeng Gao, and William~B Dolan.
  2016.
\newblock A diversity-promoting objective function for neural conversation
  models.
\newblock In \emph{Proceedings of the 2016 Conference of the North American
  Chapter of the Association for Computational Linguistics: Human Language
  Technologies}, pages 110--119.

\bibitem[{Li et~al.(2017)Li, Su, Shen, Li, Cao, and Niu}]{dialog}
Yanran Li, Hui Su, Xiaoyu Shen, Wenjie Li, Ziqiang Cao, and Shuzi Niu. 2017.
\newblock \href {https://www.aclweb.org/anthology/I17-1099} {{D}aily{D}ialog: A
  manually labelled multi-turn dialogue dataset}.
\newblock In \emph{Proceedings of the Eighth International Joint Conference on
  Natural Language Processing (Volume 1: Long Papers)}, pages 986--995, Taipei,
  Taiwan. Asian Federation of Natural Language Processing.

\bibitem[{Li et~al.(2019)Li, He, Tian, Qin, Wang, and Liu}]{hint_nat}
Zhuohan Li, Di~He, Fei Tian, Tao Qin, Liwei Wang, and Tie-Yan Liu. 2019.
\newblock \href {https://openreview.net/pdf?id=r1gGpjActQ} {Hint-based training
  for non-autoregressive translation}.
\newblock In \emph{NeuralIPS~(to appear)}.

\bibitem[{Liu et~al.(2020)Liu, Ren, Xu~Tan, Qin, Zhao, and Liu}]{liutask}
Jinglin Liu, Yi~Ren, Chen~Zhang Xu~Tan, Tao Qin, Zhou Zhao, and Tie-Yan Liu.
  2020.
\newblock Task-level curriculum learning for non-autoregressive neural machine
  translation.
\newblock \emph{AAAI}.

\bibitem[{Ma et~al.(2019)Ma, Zhou, Li, Neubig, and Hovy}]{flowseq}
Xuezhe Ma, Chunting Zhou, Xian Li, Graham Neubig, and Eduard Hovy. 2019.
\newblock \href {https://doi.org/10.18653/v1/D19-1437} {{F}low{S}eq:
  Non-autoregressive conditional sequence generation with generative flow}.
\newblock In \emph{Proceedings of the 2019 Conference on Empirical Methods in
  Natural Language Processing and the 9th International Joint Conference on
  Natural Language Processing (EMNLP-IJCNLP)}, pages 4282--4292, Hong Kong,
  China. Association for Computational Linguistics.

\bibitem[{Ott et~al.(2019)Ott, Edunov, Baevski, Fan, Gross, Ng, Grangier, and
  Auli}]{fairseq}
Myle Ott, Sergey Edunov, Alexei Baevski, Angela Fan, Sam Gross, Nathan Ng,
  David Grangier, and Michael Auli. 2019.
\newblock \href {https://doi.org/10.18653/v1/N19-4009} {fairseq: A fast,
  extensible toolkit for sequence modeling}.
\newblock In \emph{Proceedings of the 2019 Conference of the North {A}merican
  Chapter of the Association for Computational Linguistics (Demonstrations)},
  pages 48--53, Minneapolis, Minnesota. Association for Computational
  Linguistics.

\bibitem[{Papineni et~al.(2002)Papineni, Roukos, Ward, and Zhu}]{bleu}
Kishore Papineni, Salim Roukos, Todd Ward, and Wei-Jing Zhu. 2002.
\newblock \href {https://doi.org/10.3115/1073083.1073135} {{B}leu: a method for
  automatic evaluation of machine translation}.
\newblock In \emph{Proceedings of the 40th Annual Meeting of the Association
  for Computational Linguistics}, pages 311--318, Philadelphia, Pennsylvania,
  USA. Association for Computational Linguistics.

\bibitem[{Peng et~al.(2020)Peng, Ping, Song, and Zhao}]{peng2020non}
Kainan Peng, Wei Ping, Zhao Song, and Kexin Zhao. 2020.
\newblock \href {http://proceedings.mlr.press/v119/peng20a.html}
  {Non-autoregressive neural text-to-speech}.
\newblock In \emph{Proceedings of the 37th International Conference on Machine
  Learning, {ICML} 2020, 13-18 July 2020, Virtual Event}, volume 119 of
  \emph{Proceedings of Machine Learning Research}, pages 7586--7598. {PMLR}.

\bibitem[{Qian et~al.(2021{\natexlab{a}})Qian, Zhou, Bao, Wang, Qiu, Zhang, Yu,
  and Li}]{glat}
Lihua Qian, Hao Zhou, Yu~Bao, Mingxuan Wang, Lin Qiu, Weinan Zhang, Yong Yu,
  and Lei Li. 2021{\natexlab{a}}.
\newblock \href {https://arxiv.org/pdf/2008.07905.pdf} {Glancing transformer
  for non-autoregressive neural machine translation}.
\newblock In \emph{ACL}.

\bibitem[{Qian et~al.(2021{\natexlab{b}})Qian, Zhou, Zheng, Zhu, Lin, Feng,
  Cheng, Li, Wang, and Zhou}]{qian2021volctrans}
Lihua Qian, Yi~Zhou, Zaixiang Zheng, Yaoming Zhu, Zehui Lin, Jiangtao Feng,
  Shanbo Cheng, Lei Li, Mingxuan Wang, and Hao Zhou. 2021{\natexlab{b}}.
\newblock The volctrans glat system: Non-autoregressive translation meets
  wmt21.
\newblock \emph{arXiv preprint arXiv:2109.11247}.

\bibitem[{Roy et~al.(2018)Roy, Vaswani, Parmar, and Neelakantan}]{vqvae}
Aurko Roy, Ashish Vaswani, Niki Parmar, and Arvind Neelakantan. 2018.
\newblock \href {https://openreview.net/pdf?id=HkGGfhC5Y7} {Towards a better
  understanding of vector quantized autoencoders}.
\newblock \emph{arXiv}.

\bibitem[{Sennrich et~al.(2016)Sennrich, Haddow, and Birch}]{bpe}
Rico Sennrich, Barry Haddow, and Alexandra Birch. 2016.
\newblock \href {https://doi.org/10.18653/v1/P16-1162} {Neural machine
  translation of rare words with subword units}.
\newblock In \emph{Proceedings of the 54th Annual Meeting of the Association
  for Computational Linguistics (Volume 1: Long Papers)}, pages 1715--1725,
  Berlin, Germany. Association for Computational Linguistics.

\bibitem[{Shu et~al.(2019)Shu, Lee, Nakayama, and Cho}]{lv_nar}
Raphael Shu, Jason Lee, Hideki Nakayama, and Kyunghyun Cho. 2019.
\newblock \href {https://arxiv.org/pdf/1908.07181.pdf} {Latent-variable
  non-autoregressive neural machine translation with deterministic inference
  using a delta posterior}.
\newblock \emph{arXiv preprint arXiv:1908.07181}.

\bibitem[{Sun et~al.(2019)Sun, Li, Wang, He, Lin, and Deng}]{nat_crf}
Zhiqing Sun, Zhuohan Li, Haoqing Wang, Di~He, Zi~Lin, and Zhi{-}Hong Deng.
  2019.
\newblock \href
  {https://proceedings.neurips.cc/paper/2019/hash/74563ba21a90da13dacf2a73e3ddefa7-Abstract.html}
  {Fast structured decoding for sequence models}.
\newblock In \emph{Advances in Neural Information Processing Systems 32: Annual
  Conference on Neural Information Processing Systems 2019, NeurIPS 2019,
  December 8-14, 2019, Vancouver, BC, Canada}, pages 3011--3020.

\bibitem[{Sun and Yang(2020)}]{emnat}
Zhiqing Sun and Yiming Yang. 2020.
\newblock An em approach to non-autoregressive conditional sequence generation.
\newblock In \emph{International Conference on Machine Learning}, pages
  9249--9258. PMLR.

\bibitem[{Vaswani et~al.(2017)Vaswani, Shazeer, Parmar, Uszkoreit, Jones,
  Gomez, Kaiser, and Polosukhin}]{transformer}
Ashish Vaswani, Noam Shazeer, Niki Parmar, Jakob Uszkoreit, Llion Jones,
  Aidan~N. Gomez, Lukasz Kaiser, and Illia Polosukhin. 2017.
\newblock \href
  {https://proceedings.neurips.cc/paper/2017/hash/3f5ee243547dee91fbd053c1c4a845aa-Abstract.html}
  {Attention is all you need}.
\newblock In \emph{Advances in Neural Information Processing Systems 30: Annual
  Conference on Neural Information Processing Systems 2017, December 4-9, 2017,
  Long Beach, CA, {USA}}, pages 5998--6008.

\bibitem[{Wei et~al.(2019)Wei, Wang, Zhou, Lin, and Sun}]{imitate_nat}
Bingzhen Wei, Mingxuan Wang, Hao Zhou, Junyang Lin, and Xu~Sun. 2019.
\newblock \href {https://doi.org/10.18653/v1/P19-1125} {Imitation learning for
  non-autoregressive neural machine translation}.
\newblock In \emph{Proceedings of the 57th Annual Meeting of the Association
  for Computational Linguistics}, pages 1304--1312, Florence, Italy.
  Association for Computational Linguistics.

\bibitem[{Zhou et~al.(2020)Zhou, Gu, and Neubig}]{zhou2019understanding}
Chunting Zhou, Jiatao Gu, and Graham Neubig. 2020.
\newblock \href {https://openreview.net/forum?id=BygFVAEKDH} {Understanding
  knowledge distillation in non-autoregressive machine translation}.
\newblock In \emph{8th International Conference on Learning Representations,
  {ICLR} 2020, Addis Ababa, Ethiopia, April 26-30, 2020}. OpenReview.net.

\end{thebibliography}
\bibliographystyle{acl_natbib}


\appendix
\section{Details of GLAT}\label{sec:appendix_glat}
\begin{figure}[ht]
    \centering
    \begin{subfigure}[b]{0.45\textwidth}
    \includegraphics[width=1.0\linewidth]{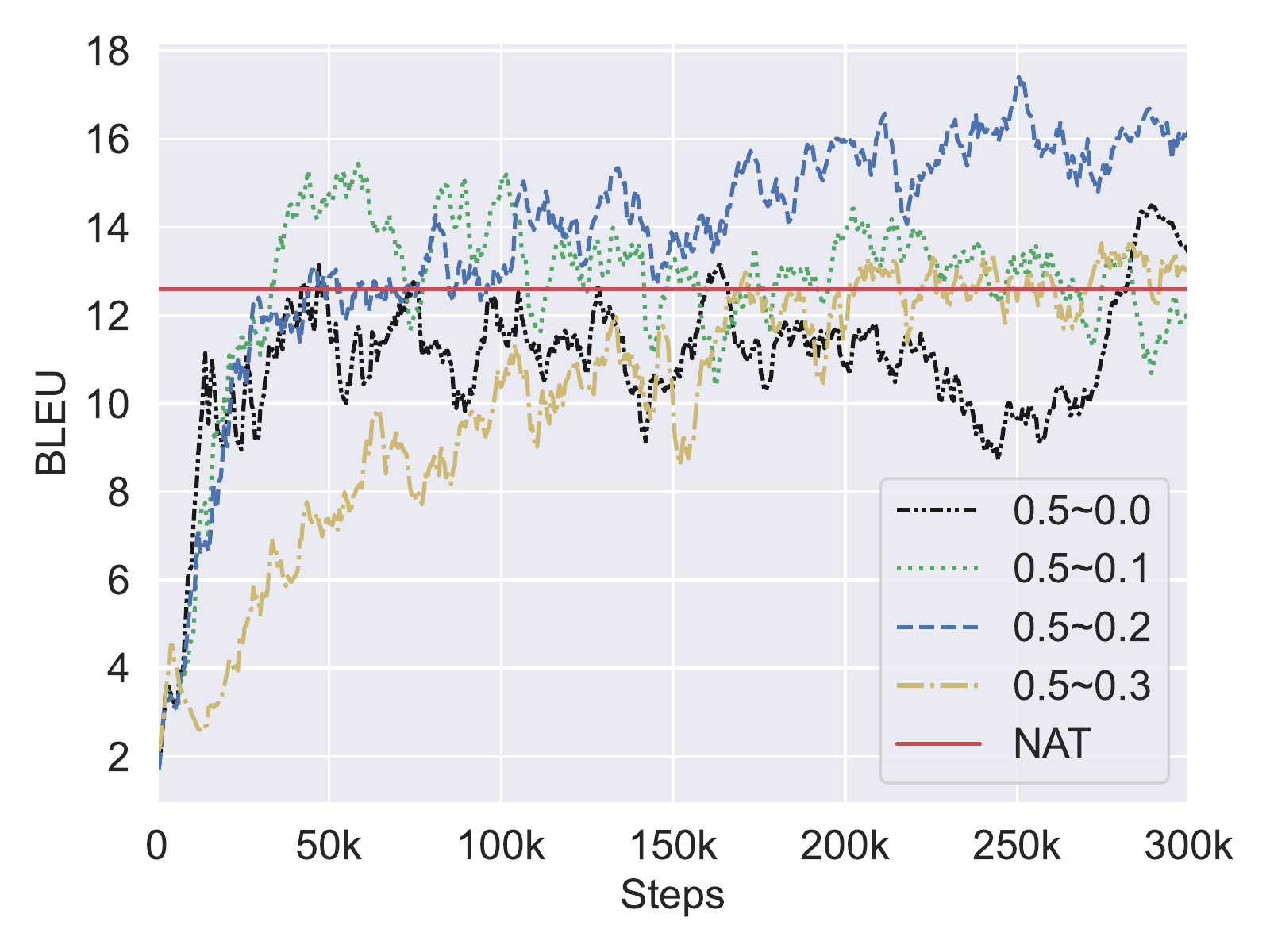}
    \caption{GLAT w/ raw data.}
    \label{fig:glat_raw}
    \end{subfigure}
    \begin{subfigure}[b]{0.45\textwidth}
    \includegraphics[width=1.0\linewidth]{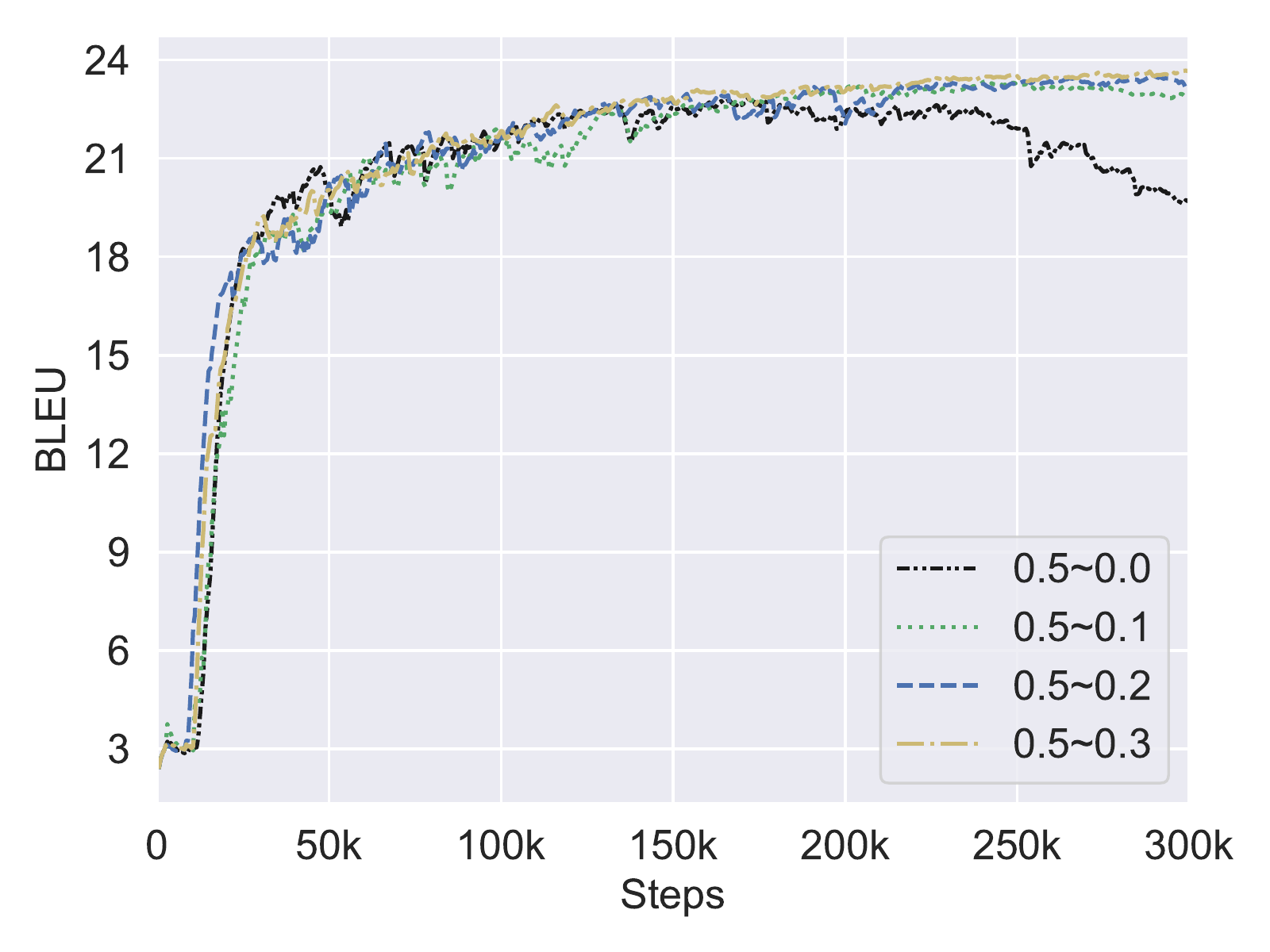}
    \caption{GLAT w/ distillation.}
    \label{fig:glat_kd}
    \end{subfigure}
\caption{BLEU score and training steps of GLAT trained with different glancing strategy (start $\to$ end ratio).}
\label{fig:GLAT_analysis}
\end{figure}

According to the performance shown in Figure~\ref{fig:glat_raw}, we can see a GLAT model will degenerate to a NAT model while using a small sampling ratio.
In such a case, introducing an autoregressive Transformer as a teacher for training the GLAT model alleviates this issue (Figure~\ref{fig:glat_kd}), indicating that the GLAT model still needs the help of knowledge distillation for alleviating multi-modality problems.

\section{Model Details of \method}\label{s:appendix_structures}

\paragraph{Decoder Inputs.} 
Following the most common practices in NAT models~\cite{imitate_nat,hint_nat}, 
we use \textit{Softcopy} mechanism for initializing the decoder inputs $\bm H=(h_1,h_2,\cdots,h_m)$:
\begin{equation}
    \begin{split}
    h_i &= \sum_{i}^{n}  \alpha_{ij}\cdot e_i,\\
    \alpha_{ij} &\propto \exp{[-(i-j\cdot\frac{n}{m})^{2}]},
    \end{split}
\end{equation}
where $\bm E=(e_1,e_2,\cdots,e_n)$ is the encoded representation of $\x=(x_1,x_2,\cdots,x_n)$, $n$ and $m$ are the length of source and target sentences, respectively.



\paragraph{Training the Latent Predictor by glancing sampling discrete latent variables.}
With the decoder input $H=h_{1:m}$ and the discretized latent variable sequence $\z=z_{1:m}$, we adopt the glancing sampling technique for training the latent predictor in the following steps: 
\begin{itemize}
    \item \textbf{Predicting} $\hat{\z}$: \method predicts the latent variable sequence with its latent predictor: $\hat{\z}\gets \F_\text{LP}(h_{1:m},e_{1:n})$.
    \item \textbf{Determining sample number $\N_\text{z}$}: Given $\z$ and $\hat{\z}$, we compute the sampling number as: 
    \begin{equation}
        \N_\text{z} = \tau\cdot \operatorname{Hamming}(\z,\hat{\z})
    \end{equation}
    where $\tau$ is the sampling ratio decreasing in the training steps, and we use $\operatorname{Hamming}$ distance~\cite{hamming1950error} for measuring the prediction quality.
    \item \textbf{Sampling observed latent variables $\obs{\z}$}: Given discretized latent variable sequence $\z$ and sample number $\N_\text{z}$, we obtain $\obs{\z}$ by random selecting $N_\text{z}$ elements from $\z$.  
    \item \textbf{Re-constructing inputs $H_\text{LP}$}: We construct $H_\text{LP}$ by position-wise replacing the decoder input $h_{1:m}$ with $\obs{\z}$.
    \item \textbf{Updating Latent Predictor}: With the $H_\text{LP}$ as inputs, we train the latent predictor to predict the unobserved references $\uno{\z}$.
\end{itemize}

\paragraph{Training the Mix. Decoder with sampled discrete latent variables.}
Training of Mix. Decoder is largely follow the ~\citet{glat}, except using extra latent variables as inputs.
With the input $H=h_{1:m}$, the reference sentence $\y$, and the sampled latent variables $\obs{\z}$, we train Mix. Decoder in the following steps: 
\begin{itemize}
    \item \textbf{Predicting $\hat{\y}$}: \method predicts the target sentences: $\hat{\y}\gets \F_\text{DEC}(\obs{\z},h_{1:m},e_{1:n})$.
    \item \textbf{Determining sample number $N_\text{y}$}: Given $\y$ and $\hat{\y}$, we compute the sampling number $N_\text{y} = \tau\cdot \operatorname{Hamming}(\y,\hat{\y})$.
    \item \textbf{Sampling target tokens $\obs{\y}$}: We obtain the glancing reference $\obs{\y}$ by random selecting $N_\text{y}$ tokens from reference sequence $\y$.
    \item \textbf{Re-constructing inputs $H_\text{DEC}$}: $H_\text{DEC}$ is constructed by position-wise replacing the decoder input $H$ with embedding of $\obs{\y}$.
    \item \textbf{Updating Mix. Decoder}: We then train the Mix. Decoder to predict the unobserved references $\uno{\y}$, with the $H_\text{DEC}$ and $\obs{\z}$ as inputs.
\end{itemize}

\end{document}